\documentclass[11pt]{article}

\usepackage[preprint]{acl}

\usepackage{times}
\usepackage{latexsym}

\usepackage[T1]{fontenc}

\usepackage[utf8]{inputenc}

\usepackage{microtype}

\usepackage{inconsolata}

\usepackage{graphicx}
\usepackage{booktabs}
\usepackage{multirow}
\usepackage{pifont}
\usepackage{url}
\title{\Name: Stage-wise and Orientation-specific Benchmarking for Large Language Models in E-commerce}

\author{
 \textbf{Kaiyan Zhao$^\dagger$\textsuperscript{1}},
 \textbf{Zijie Meng$^\dagger$\textsuperscript{2}},
 \textbf{Zheyong Xie\textsuperscript{3}},
 \textbf{Jin Duan\textsuperscript{3}},
 \textbf{Yao Hu\textsuperscript{3}}, \\
 \textbf{Zuozhu Liu\textsuperscript{2}},
 \textbf{Shaosheng Cao$^\ddagger$\textsuperscript{3}
 }
\\
 \textsuperscript{1}The University of Tokyo,
 \textsuperscript{2}Zhejiang University,
 \textsuperscript{3}Xiaohongshu Inc.
\\
\normalfont{\fontsize{11pt}{12pt}\selectfont {\fontfamily{qcr}\selectfont caoshaosheng@xiaohongshu.com}} \\
}

\newcommand{\Name}{\textcolor{black}{EComStage}}

\begin{document}
\maketitle
\begin{abstract}
Large Language Model (LLM)-based agents are increasingly deployed in e-commerce applications to assist customer services in tasks such as product inquiries, recommendations, and order management. Existing benchmarks primarily evaluate whether these agents successfully complete the final task, overlooking the intermediate reasoning stages that are crucial for effective decision-making. To address this gap, we propose \Name, a unified benchmark for evaluating agent-capable LLMs across the comprehensive stage-wise reasoning process: Perception (understanding user intent), Planning (formulating an action plan), and Action (executing the decision). \Name~evaluates LLMs through seven separate representative tasks spanning diverse e-commerce scenarios, with all samples human-annotated and quality-checked. Unlike prior benchmarks that focus only on customer-oriented interactions, \Name~also evaluates merchant-oriented scenarios, including promotion management, content review, and operational support relevant to real-world applications. We evaluate a wide range of over 30 LLMs, spanning from 1B to over 200B parameters, including open-source models and closed-source APIs, revealing stage/orientation-specific strengths and weaknesses. Our results provide fine-grained, actionable insights for designing and optimizing LLM-based agents in real-world e-commerce settings.\footnote{$^\dagger$Equal contribution. $^\ddagger$Corresponding author. Codes and data: \url{https://github.com/KYuuto1006/EComStage}}
\end{abstract}

\section{Introduction}
Large Language Model (LLM)-based agents have demonstrated remarkable reasoning and decision-making abilities across a wide range of complex tasks \citep{yang2023autogptonlinedecisionmaking, luo2025largelanguagemodelagent}. They have been widely adopted in real-world applications, such as code generation \citep{dong2025surveycodegenerationllmbased} and gaming \citep{hu2025gaminglargelanguagemodelbased}.

Among these domains, e-commerce stands out as one of the most promising and challenging areas for LLM-based agents. These agents are expected to enhance customers' online shopping experiences by handling diverse and goal-oriented interactions~\cite{zeng2025citespeakenhancingcontextresponse}. To achieve this, they must interpret user intents accurately while dynamically invoking a variety of tools such as databases and APIs to browse products, manage transactions, and resolve service issues~\cite{yao2024taubenchbenchmarktoolagentuserinteraction, EC-Guide}. 

The practical value of LLM-based agents in e-commerce lies in their ability to automate large portions of customer interaction workflows. Major online retailers and service providers have begun integrating such agents into their platforms to assist with product search, order management, and customer support, demonstrating their growing practical impact~\cite{yao2022webshop, Cheng2024amazon, zeng2025citespeakenhancingcontextresponse}.

Given their growing importance in e-commerce applications, accurately evaluating the performance of LLM-based agents has become an emerging research focus. Several e-commerce–specific benchmarks have been proposed. For example, $\tau$-bench \cite{yao2024taubenchbenchmarktoolagentuserinteraction} introduces evaluation scenarios in the \textit{retail} and \textit{airline} domains to assess agents’ capabilities in customer service assistance. ECom-Bench \cite{wang2025ecombenchllmagentresolve} emphasizes the evaluation of agents' multimodal reasoning abilities, while Mix-Ecom \cite{zhou2025mixecommixedtypeecommercedialogues} focuses on mixed-type e-commerce dialogues, such as multi-intent conversations involving refund requests followed by product recommendations.

\begin{figure*}[ht]
  \centering
  \includegraphics[width=0.9\textwidth]{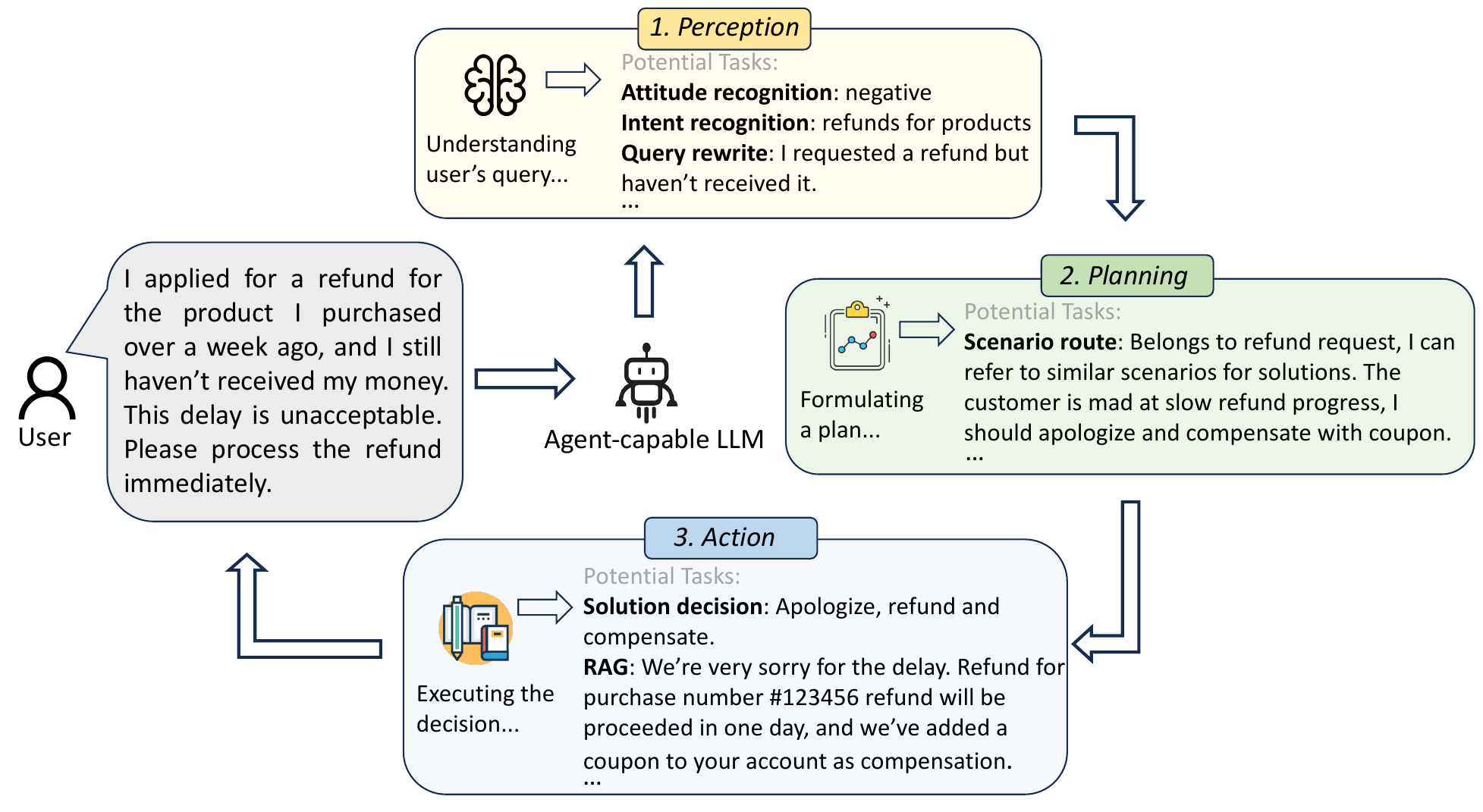}
  \caption{Stage-wise reasoning process of an LLM-based agent when handling an e-commerce requirement.}
  \label{fig1}

\end{figure*}

However, existing e-commerce benchmarks share one common limitation: they primarily evaluate whether the agent successfully completes the final task, overlooking the intermediate stages that are crucial for effective decision-making. Before reaching a final decision, the backbone LLMs typically engage in a multi-stage reasoning process, and these stages play a vital role in overall performance. As illustrated in Figure~\ref{fig1}, when given a single e-commerce requirement, we can decompose the agent's reasoning process into three stages: (1) Perception: recognizing and summarizing the true intent of the customer; (2) Planning: reasoning and formulating a plan of action; and (3) Action: executing the appropriate decision to assist the customer. Each stage contributes a distinct aspect of reasoning, and together they enable the agent to achieve the final goal. Throughout this process, the agent must continuously interpret the customer’s current intent to decide the most appropriate next step. Unfortunately, existing e-commerce benchmarks lack a comprehensive evaluation of these reasoning stages, focusing instead solely on the final task success rates.

Moreover, understanding how agent-capable LLMs perform in these stages is crucial for improving real-world applications. For instance, failures in the Perception stage often lead to misunderstanding customer intent, causing irrelevant responses or poor recommendations; weaknesses in Planning can result in inefficient tool use or inconsistent task strategies; and errors in Action directly affect customer satisfaction and business outcomes. Therefore, evaluating LLMs that serve as the backbone of e-commerce agents across these stages provides actionable insights for improving user experience in real e-commerce applications.

While some prior work has emphasized the importance of these intermediate reasoning stages~\cite{liuagentbench, pipelineisa2024experimentalevaluationmachinelearning}, and others have considered Perception as a distinct step~\cite{huang2025memorbplugandplayverbalreinforcementmemory}, an integrated evaluation framework encompassing all three stages remains missing, as annotating and evaluating intermediate stages remains challenging and costly. 

To this end, we propose \Name, a comprehensive benchmark for evaluating LLMs that serve as the backbone of e-commerce agents across the full three-stage reasoning process. Unlike previous benchmarks that focus solely on end-to-end task outcomes, \Name~decomposes the e-commerce reasoning process into separate, stage-specific tasks, each designed to evaluate a distinct capability of the agent-capable LLM. Specifically, \Name~ introduces seven representative tasks covering Perception, Planning, and Action abilities under diverse e-commerce settings, such as product inquiry, recommendation, and refund handling. This design enables a more fine-grained and interpretable evaluation, helping both researchers and practitioners identify stage-specific weaknesses and optimize agent behavior for real-world applications. To ensure reliability and alignment with real-world settings, all samples are human-annotated and quality-checked by professional annotators with e-commerce experience.  In addition to traditional customer-service tasks, \Name~explicitly includes \textbf{merchant-oriented} operational scenarios, such as promotion management, content review, and refund handling for advertisers. This extension allows evaluation of LLM-based agents in supporting both customers and merchants, reflecting the full spectrum of real-world e-commerce interactions.

We evaluate a diverse set of over 30 agent-capable LLMs on \Name, ranging from standard open-source LLMs such as LLaMA3 \cite{grattafiori2024llama3herdmodels} and Qwen3 \cite{yang2025qwen3technicalreport} series to closed-source APIs including GPT-4o \cite{openai2024gpt4ocard} and Gemini 2.5 Pro \cite{comanici2025gemini25pushingfrontier}, with model sizes spanning from 1B to over 200B parameters. By benchmarking this wide range of models, we aim to demonstrate both the generalizability of \Name~and the stage/orientation-specific strengths and weaknesses of current LLM-based agents in realistic e-commerce scenarios. Through extensive evaluation, our experiments reveal that no single model consistently excels across all stages (Perception, Planning, Action) or orientations (merchant-oriented, customer-oriented). This finding underscores the importance of a stage-wise and orientation-aware benchmark.

In summary, our paper makes the following contributions:
\begin{itemize}
    
    \item We propose \Name, which provides a stage-wise evaluation framework for agent-capable LLMs in e-commerce, highlighting intermediate reasoning that is often overlooked in existing benchmarks.
    \item \Name~introduces seven representative tasks covering diverse real-world scenarios including both customer and merchant orientation, with all samples human-annotated and quality-checked, enabling fine-grained, stage-specific evaluation.
    \item By analyzing multiple models of varying sizes and capabilities, our benchmark offers valuable, actionable insights into agent performance, guiding both research and practical deployment in real-world e-commerce systems. 
\end{itemize}

\section{Related Works}
\begin{table*}[t]
\centering
\resizebox{0.99\textwidth}{!}{
\begin{tabular}{lcccccc}
\toprule
Benchmark            & Scale &Tested Models   & Customer-oriented & Merchant-oriented & Stage-wise Evaluation & Availability \\ 
\midrule
ECom-Bench \cite{wang2025ecombenchllmagentresolve}           & 53$^\dagger$ & 7 & \ding{51}            &      \ding{55}       &          \ding{55}             & \ding{51}    \\
Mix-Ecom \cite{zhou2025mixecommixedtypeecommercedialogues}            & 4799    & 6  &   \ding{51}         &      \ding{55}       &       \ding{55}                & Not yet      \\
\Name~(ours) & 4804    & 33    &  \ding{51}        &     \ding{51}        &     \ding{51}                  &  \ding{51}      \\ \bottomrule
\end{tabular}}
\caption{Comparison of Existing E-Commerce Benchmarks and \Name. $^\dagger$Based on the released ECom-Bench repository, we observe only 53 samples.}
\label{benchmark_comparison}
\end{table*}

\paragraph{LLM-based Agents in E-commerce} 
LLM-based agents have recently gained widespread attention for their potential to automate complex, multi-step workflows across different domains \cite{xi2023risepotentiallargelanguage, guo2024largelanguagemodelbasedmultiagent, luo2025largelanguagemodelagent}. By integrating the reasoning capabilities from LLMs, these agents can interact with users, retrieve external information, and execute tasks autonomously. 
In the e-commerce industry, LLM-based agents hold particular significance as they directly influence customer experience, operational efficiency, and seller profitability through intelligent service automation and personalized interactions \cite{zeng2025citespeakenhancingcontextresponse}.

\paragraph{Early E-commerce Datasets} 
Despite this growing importance, evaluation of LLM-based agents' performance in e-commerce scenarios remains underexplored.
Early e-commerce–related datasets often originate as subsets or adaptations of multi-domain task-oriented datasets, e.g., MultiWOZ~\cite{budzianowski-etal-2018-multiwoz}, which primarily focuses on limited scenarios, such as booking and reservation rather than real e-commerce operations.
Later e-commerce datasets begin to include domain-specific elements such as product recommendation \cite{jia-etal-2022-econvrec, U-NEED} and content detection \cite{xu2025evademultimodalbenchmarkevasive}, yet they remain limited in scope, often modeling only one aspect of the shopping experience rather than  the diverse interactions found in real platforms.

\paragraph{E-commerce Benchmarks}
Recent works have begun to address these limitations with e-commerce–specific benchmarks. $\tau$-Bench~\cite{yao2024taubenchbenchmarktoolagentuserinteraction} simulates realistic tool–agent–user interactions in the retail and airline domains, while
ECom-Bench~\cite{wang2025ecombenchllmagentresolve} assesses agents’ capabilities in resolving customer support issues, including multimodal reasoning and interaction with structured data.
Mix-Ecom~\cite{zhou2025mixecommixedtypeecommercedialogues} introduces mixed-type e-commerce dialogues, covering pre-sales, logistics, and after-sales scenarios. It evaluates agents’ ability to handle complex multi-turn interactions.
While these benchmarks have advanced evaluation in e-commerce, they mainly focus on customer-oriented tasks and lack fine-grained analysis of intermediate reasoning stages. EComScriptBench~\cite{wang-etal-2025-ecomscriptbench} introduces step-wise evaluation in script planning, but its scope is restricted and does not cover broader agent behavior.

To ensure safe and reliable deployment, it is essential to evaluate not only whether agents complete a task, but also how effectively they reason through intermediate stages. Our \Name~enables a more comprehensive evaluation of the capabilities of LLMs that serve as the backbone of e-commerce agents, capturing performance from different stakeholder perspectives and supporting both customer- and merchant-oriented scenarios. We summarize the main differences between existing e-commerce benchmarks and \Name~in Table~\ref{benchmark_comparison}.

\section{\Name}

\subsection{Dataset Construction}

\begin{figure}[t]
  \centering
  \includegraphics[width=0.49\textwidth]{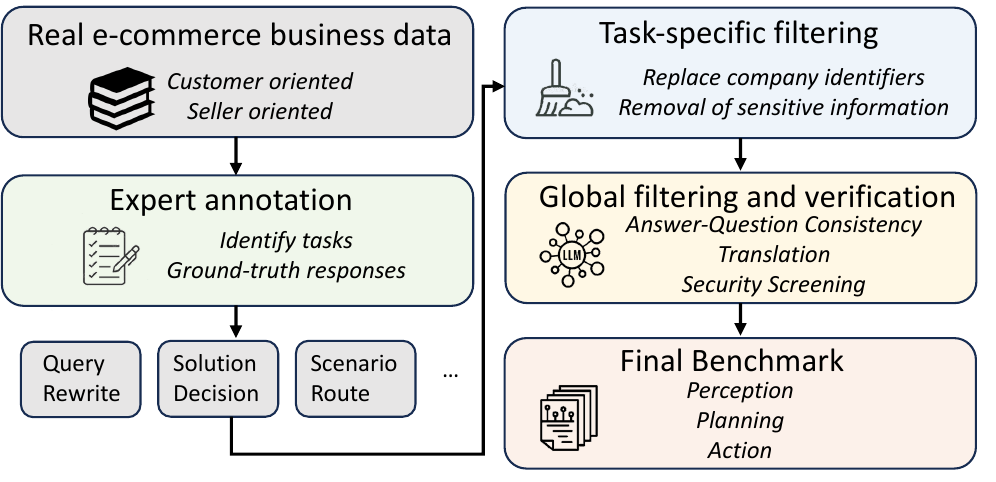}
  \caption{Pipeline for dataset construction.}
  \label{figpipe}

\end{figure}

To ensure both realism and safety, we construct \Name~through a multi-stage pipeline that integrates real-world e-commerce data collection, expert annotation, and rigorous multi-level filtering as shown in Figure~\ref{figpipe}. 

\subsubsection{Data Collection}
We first collect business data from real e-commerce service scenarios, covering both customer and merchant orientations (e.g., customer inquiries, promotion management, advertising content review). These data are drawn from authentic operational contexts to ensure that each task reflects realistic e-commerce interaction patterns. From this data, we define seven representative tasks, chosen because they are the most frequently encountered and critical in real-world e-commerce workflows, ensuring that our benchmark evaluates capabilities relevant to practical applications.

\subsubsection{Human Annotation}
All collected samples are manually annotated by professional annotators with e-commerce experience. Annotators identify the task types, understand user intents, determine correct actions, and write corresponding ground-truth responses following well-defined task guidelines on an online platform. This process ensures high-quality supervision and domain alignment for each task. Guidelines for human annotators are provided in Figure~\ref{guideline}, Appendix.

\subsubsection{Task-specific Filtering}
Data from each task undergoes an independent filtering phase to eliminate legal, security, or privacy risks through the following steps:
(a) Removal of company identifiers and overly detailed internal operational procedures; (b) Rewriting or merging of text involving product rules or classification results to maintain generality; (c) Anonymization of sensitive user information, including names, phone numbers, addresses, ID numbers, order details, and product metadata; and (d) For samples containing images, all identifiable personal information is blurred or replaced.

\subsubsection{Global Filtering and Verification}

After task-level cleaning, all tasks are processed through a unified multi-step quality and safety pipeline: (a) Answer–Question Consistency Check (LLM): An LLM evaluates whether each answer appropriately corresponds to its question; (b) Translation: All data are translated from Chinese into English to enhance accessibility and standardization; (c) Re-Consistency Check (LLM): The translated samples are re-evaluated for semantic consistency between question and answer; and (d) Security Screening: Both automated safety APIs and LLM-based evaluation again assess potential sensitive or risky content in both questions and answers.

All LLM-based filtering in the pipeline are performed using Qwen3-235B-A22B\footnote{\url{https://huggingface.co/Qwen/Qwen3-235B-A22B}}, selected for its high reasoning capability and multilingual understanding.

This multi-stage construction ensures that \Name~balances authenticity, safety, and reproducibility, making it suitable for both academic research and industrial deployment. Detailed prompts for dataset construction are provided in Appendix~\ref{prompts}.

\subsection{Tasks}
\begin{table}[t]
\centering
\resizebox{0.49\textwidth}{!}{
\begin{tabular}{cccc}
\toprule
\multicolumn{1}{c|}{Capability categories}                            & \multicolumn{1}{c|}{Tasks} & \multicolumn{1}{c|}{Orientation}                                   & Instances \\ 
\midrule
\multicolumn{1}{c|}{\multirow{4}{*}{Perception}} & \multicolumn{1}{l|}{Query Rewrite}    & \multicolumn{1}{c|}{Customer}   & 233       \\
\multicolumn{1}{c|}{}                            & \multicolumn{1}{l|}{Attitude Classification}  & \multicolumn{1}{c|}{Merchant}      & 424       \\
\multicolumn{1}{c|}{}                            & \multicolumn{1}{l|}{Query Match}    & \multicolumn{1}{c|}{Customer}     & 1927      \\
\multicolumn{1}{c|}{}                            & \multicolumn{1}{l|}{Intent Recognition} & \multicolumn{1}{c|}{Customer} & 1367      \\ 
\midrule
\multicolumn{1}{c|}{Planning}                    & \multicolumn{1}{l|}{Scenario Route} & \multicolumn{1}{c|}{Merchant}     & 164       \\ 
\midrule
\multicolumn{1}{c|}{\multirow{2}{*}{Action}}     & \multicolumn{1}{l|}{Solution Decision}  & \multicolumn{1}{c|}{Customer} & 487       \\
\multicolumn{1}{l|}{}                            & \multicolumn{1}{l|}{RAG-QA}         & \multicolumn{1}{c|}{Both}     & 202       \\ 
\midrule
\multicolumn{3}{c|}{Total instances}                                                        & 4804      \\ 
\bottomrule
\end{tabular}}
\caption{Statistics of \Name.}
\label{statistics}
\end{table}

We categorize our benchmark tasks according to the three-stage reasoning framework introduced in former sections: Perception, Planning, and Action. Each task is designed to evaluate stage-specific abilities of agent-capable LLMs in realistic e-commerce scenarios.

\subsubsection{Perception}
Tasks in this category measure the model’s ability to understand user intent and relevant context from both customer and merchant orientations.
\begin{itemize}
    \item \textbf{Query Rewrite} requires the agent to rewrite the user’s last utterance based on the conversation history, making it clearer and more complete while preserving the original meaning.
    \item \textbf{Attitude Classification} asks the agent to identify whether the merchant’s current message contains a negative attitude based on chat history.
    \item \textbf{Query Match} evaluates the agent’s ability to match a customer query to the most relevant entry in a provided question lists.
    \item \textbf{Intent Recognition} asks the agent to identify the underlying intention of a customer message, including refund requests, complaints, product inquiries, and other issues.
\end{itemize}

\subsubsection{Planning} 
This stage assesses the model’s reasoning and action formulation capabilities. 
\begin{itemize}
    \item \textbf{Scenario Route} requires the agent to determine the correct merchant-oriented workflow based on historical chat records given pre-defined scenarios, such as content rejection, promotion complaints, or refund requests. 
\end{itemize}
This evaluates the agent’s ability to reason over multi-turn conversations and select the appropriate operational path.

\subsubsection{Action}
Action tasks test the model’s decision-making and response execution. 
\begin{itemize}
    \item \textbf{Solution Decision} requires the agent to generate the most appropriate response or solution for a customer query, referencing relevant response choices. 
    \item \textbf{RAG-QA} evaluates the agent’s ability to retrieve relevant reference knowledge and generate a context-aware, accurate answer, supporting both customer and merchant scenarios.
\end{itemize}

The statistics of \Name, including task distribution, categories, and customer/merchant orientations after filtering, are summarized in Table~\ref{statistics}. Note that for Query Match and Solution Decision, human annotators provide pre-defined question lists and response options tailored to each data sample. Specifically, our Planning set contains only 164 samples, but each sample spans multiple merchant scenarios, making it highly informative despite the smaller quantity. Overall, the benchmark includes five close-ended tasks and two open-ended generation tasks. 
The example prompts for each task are provided in Appendix~\ref{examples}.

\section{Experiments}
\subsection{Experimental Settings}
\subsubsection{Evaluated Models}
We evaluate a diverse set of agent-capable LLMs on \Name. The selected models span multiple families, including general-purpose language models, instruction-tuned variants, MoE variants, and multimodal ones, reflecting the diversity of current research and industrial systems. We list the evaluated models as follows:
\begin{itemize}
    \item Closed-source APIs: GPT-4o \cite{openai2024gpt4ocard}, Gemini 2.5-Pro \cite{comanici2025gemini25pushingfrontier}, Claude 3.7 \cite{Claude3S}, Claude Sonnet 4 \cite{Claude4}.
    \item Open-source LLMs: LLaMA3.2 (3B, 11B), LLaMA3.3 (70B) \cite{grattafiori2024llama3herdmodels}, Qwen2.5-Instruct (3B, 7B, 14B, 32B, 72B) \cite{qwen2025qwen25technicalreport}, Qwen3 (1.7B, 4B, 4B-Instruct, 8B, 14B, 30B-A3B, 30B-A3B-Instruct, 32B, 235B-A22B, 235B-A22B-Instruct) \cite{yang2025qwen3technicalreport}, Phi-4-mini \cite{abdin2024phi4technicalreport}, GLM4 (9B, 32B) \cite{glm2024chatglm}, InternVL3 (8B, 14B) \cite{zhu2025internvl3exploringadvancedtraining}, MiniCPM-8B \cite{minicpm4}, DeepSeek-V3 \cite{deepseekai2025deepseekv3technicalreport}, DeepSeek-R1 \cite{deepseekai2025deepseekr1incentivizingreasoningcapability}, dots.llm1.inst \cite{huo2025dotsllm1technicalreport}, gpt-oss-120B \cite{openai2025gptoss120bgptoss20bmodel}.
\end{itemize}
We choose this broad collection of models to assess the impact of model scale and cover multiple model families and architectures relevant for real-world e-commerce scenarios.

\begin{table*}[ht]
\centering
\resizebox{0.99\textwidth}{!}{
\begin{tabular}{l|cccc|c|cc|c}
\toprule
\multicolumn{1}{c|}{\multirow{2.5}{*}{Models}}                                                    & \multicolumn{4}{c|}{Perception}                                                                                                                             & \multicolumn{1}{c|}{Planning}       & \multicolumn{2}{c|}{Action}                                          & \multirow{2.5}{*}{Avg.} \\ 
\cmidrule(lr){2-5} \cmidrule(lr){6-6} \cmidrule(lr){7-8}
\multicolumn{1}{c|}{}                                                                           & \multicolumn{1}{c}{Query Rewrite} & \multicolumn{1}{c}{Attitude Classification} & \multicolumn{1}{c}{Query Match} & \multicolumn{1}{c|}{Intent Recognition} & \multicolumn{1}{c|}{Scenario Route} & \multicolumn{1}{c}{Solution Decision} & \multicolumn{1}{c|}{RAG-QA} &                       \\ \midrule
\multicolumn{9}{c}{\textit{Closed-source APIs}}                                                                                                   \\ \midrule
GPT4o &  80.70 &  \underline{79.95} &  97.77 &  \underline{90.07} &  \underline{85.98} &  \underline{79.47} &  68.24 &  83.17       \\
Gemini2.5-Pro  &  81.42 &  78.30 &  \textbf{98.75} &  89.03 &  83.54 &  \textbf{87.06} &  \underline{69.97} &  \underline{84.01}      \\
Claude-3.7 &  \textbf{81.99} &  77.59 &  98.39 &  88.00 &  82.32 &  75.56 &  69.38 &  81.89       \\
Claude Sonnet 4    &  \underline{81.81} &  \textbf{83.49} &  \underline{98.44} &  \textbf{90.42} &  \textbf{88.41} &  74.74 &  \textbf{72.13} &  \textbf{84.21}   \\\midrule
\multicolumn{9}{c}{\textit{Open-source Models \textless 7B}}                                       \\ \midrule
Qwen2.5-1.5B-Instruct    &  74.72 &  64.62 &  93.93 &  43.96 &  61.59 &  \underline{78.85} &  64.85 &  68.93        \\
Qwen2.5-3B-Instruct   &  76.19 &  70.99 &  95.12 &  68.98 &  53.66 &  77.62 &  64.72 &  72.47       \\
Qwen3-1.7B    &  69.33 &  76.42 &  91.28 &  63.42 &  60.98 &  53.39 &  66.57 &  68.77       \\
Qwen3-4B   &  73.26 &  \underline{78.07} &  \textbf{97.09} &  \underline{80.83} &  71.95 &  78.23 &  \underline{67.47} &  \underline{78.13}        \\
Qwen3-4B-Instruct &  \textbf{79.97} &  \textbf{82.78} &  \underline{96.94} &  \textbf{84.20} &  \textbf{78.66} &  \textbf{84.80} &  \textbf{68.45} &  \textbf{82.26} \\
Llama3.2-3B   &  \underline{76.99} &  69.58 &  89.93 &  44.48 &  \underline{77.44} &  15.61 &  63.73 &  62.54      \\
Phi-4-mini (4B)   &  76.53 &  71.70 &  91.49 &  74.62 &  75.61 &  61.60 &  62.23 &  73.40       \\\midrule
\multicolumn{9}{c}{\textit{Open-source Models \textless 30B}}     \\ \midrule
Qwen2.5-7B-Instruct      &  75.69 &  78.07 &  96.89 &  80.61 &  75.00 &  72.48 &  66.89 &  77.95        \\
Qwen2.5-14B-Instruct    &  79.06 &  81.60 &  \underline{97.98} &  85.59 &  \textbf{82.93} &  80.90 &  66.24 &  \underline{82.04}       \\
Qwen3-8B    &  75.40 &  79.25 &  97.30 &  83.61 &  78.05 &  \textbf{85.01} &  \textbf{68.48} &  81.01      \\
Qwen3-14B   &  77.74 &  \underline{82.08} &  \textbf{98.75} &  \textbf{86.98} &  \underline{81.10} &  77.21 &  \underline{68.04} &  81.70       \\
Llama3.2-11B            &  78.73 &  74.29 &  94.97 &  75.64 &  73.78 &  67.35 &  64.75 &  75.64      \\
GLM4-9B  &  \underline{80.39} &  38.44 &  95.38 &  77.32 &  80.49 &  72.90 &  67.17 &  73.16   \\
InternVL3-8B  &  77.73 &  81.60 &  97.87 &  84.64 &  78.05 &  68.38 &  66.09 &  79.19 \\
InternVL3-14B  &  \textbf{80.82} &  \textbf{85.85} &  97.77 &  \underline{86.76} &  78.66 &  \underline{81.11} &  65.86 &  \textbf{82.40}\\
MiniCPM-8B  &  78.31 &  80.19 &  94.29 &  62.37 &  57.93 &  72.28 &  64.19 &  72.79 \\\midrule
\multicolumn{9}{c}{\textit{Open-source Models $\geq$ 30B}}    \\ \midrule
Qwen2.5-32B-Instruct  &  79.74 &  79.95 &  98.50 &  88.15 &  85.37 &  76.59 &  65.88 &  82.03      \\
Qwen2.5-72B-Instruct  &  80.28 &  83.02 &  98.70 &  \textbf{89.83} &  \textbf{89.02} &  87.06 &  67.06 &  \underline{85.00}  \\
Qwen3-32B  &  79.37 &  84.43 &  98.44 &  87.42 &  75.61 &  73.51 &  68.73 &  81.07       \\
Qwen3-30B-A3B  &  70.07 &  82.08 &  97.98 &  85.30 &  \textbf{89.02} &  81.72 &  68.21 &  82.05       \\
Qwen3-30B-A3B-Instruct &  78.53 &  78.30 &  97.92 &  86.32 &  82.32 &  \textbf{94.25} &  67.75 &  83.63 \\
Qwen3-235B-A22B  &  78.83 &  87.26 &  \textbf{98.81} &  86.91 &  82.32 &  87.47 &  69.41 &  84.43     \\
Qwen3-235B-A22B-Instruct &  \underline{81.04} &  87.26 &  \underline{98.75} &  87.78 &  84.76 &  \underline{89.94} &  \underline{69.76} &  \textbf{85.61} \\
DeepSeek-V3  &  \textbf{82.49} &  \textbf{88.68} &  98.34 &  88.59 &  85.37 &  77.00 &  69.17 &  84.23       \\
DeepSeek-R1  &  79.52 &  83.02 &  98.70 &  \underline{89.61} &  85.98 &  66.74 &  \textbf{71.65} &  82.17        \\
GLM4-32B  &  78.31 &  83.73 &  97.56 &  84.13 &  \underline{87.80} &  77.00 &  69.87 &  82.63       \\
LLama3.3-70B  &  80.26 &  79.01 &  98.08 &  86.83 &  84.76 &  70.23 &  66.63 &  80.83       \\
dots.llm1.inst &  79.61 &  \underline{87.74} &  98.08 &  71.40 &  70.73 &  74.13 &  64.92 &  78.09      \\
GPT-OSS-120B  &  78.38 &  73.58 &  98.39 &  88.59 &  77.44 &  78.23 &  70.04 &  80.66
   \\ 
\bottomrule                                       
\end{tabular}}
\caption{Main evaluation results. The best results are highlighted in bold, and the second-best results are underlined.}
\label{main_exp}
\end{table*}

\subsubsection{Evaluation Metrics}
For close-ended tasks, we report accuracy, while for open-ended generation tasks, we use cosine similarity to measure alignment with reference answers provided by human annotators. Specifically, we use Qwen3-Embedding-8B\footnote{\url{https://huggingface.co/Qwen/Qwen3-Embedding-8B}} \cite{zhang2025qwen3embeddingadvancingtext} to convert the generated texts and reference answers into embeddings of 4096 dimensions and then calculate the cosine similarity between them.

\subsubsection{Implementation Details}
For all experiments, we set the batch size to 32 and limit the model input length to 4,096 tokens. During generation, we use a low temperature of 0.1 and a top-p sampling threshold of 0.001 to encourage deterministic outputs. To reduce repetitive responses, we apply a repetition penalty of 1.05. The maximum number of tokens generated per query is capped at 512. These settings are applied consistently across all evaluated models to provide a fair comparison and simulate practical deployment scenarios. All experiments are conducted on 8 NVIDIA H800 GPUs with a single run. 

\subsection{Main Experimental Results}

We perform systematic evaluation of LLM-based agents’ abilities at each reasoning stage, from Perception and Planning to Action.
The results across the diverse set of evaluated models are summarized in Table~\ref{main_exp}. 

\subsubsection{Closed-source APIs}
We first evaluate closed-source APIs, as shown at the top of Table~\ref{main_exp}. These models demonstrate strong overall performance across all tasks. Among them, Claude Sonnet 4 achieves the best average score (84.21), followed closely by Gemini 2.5-Pro (84.01). This finding is consistent with prior observations \cite{zhou2025mixecommixedtypeecommercedialogues}, where both models exhibit superior performance in e-commerce scenarios compared to GPT-4o. 
While all models perform consistently well on classification-style tasks such as Query Match and Intent Recognition, their performance varies more on merchant-oriented tasks like Attitude Classification and Scenario Route, highlighting the limitations of current evaluation practices. Claude Sonnet 4 shows stronger capability in handling merchant-oriented tasks, likely due to its optimization for complex reasoning and tool use~\cite{Claude4}, making it a strong baseline for agent-oriented evaluation. In contrast, Gemini 2.5-Pro achieves the highest accuracy in Solution Decision, benefiting from its enhanced long-term planning ability \cite{comanici2025gemini25pushingfrontier}.

\subsubsection{Open-source Models less than 7B}
We next evaluate open-source models with fewer than 7B parameters. Among them, Qwen3-4B-Instruct achieves the highest overall score (82.26), showing strong generalization across both customer- and merchant-oriented tasks. It performs particularly well on Attitude Classification and Scenario Route, indicating improved instruction-following and reasoning capabilities. This strong performance aligns with its results on $\tau$-Bench \cite{yao2024taubenchbenchmarktoolagentuserinteraction}, suggesting that the model’s recent instruction tuning and alignment optimization substantially improve its robustness and domain adaptability in complex reasoning scenarios \cite{yang2025qwen3technicalreport}.
In contrast, smaller models such as Qwen2.5-1.5B-Instruct and Llama3.2-3B struggle on complex reasoning and decision-oriented tasks like Intent Recognition and Solution Decision, partly because they belong to relatively earlier generations of LLMs \cite{zhao2025surveylargelanguagemodels}, developed prior to recent advances in multi-turn instruction tuning and reasoning optimization. Despite the smaller size, 4B models also demonstrate balanced accuracy across tasks, showing potential for lightweight deployment.

\subsubsection{Open-source Models less than 30B}
Moving to mid-sized models (below 30B parameters), we observe that their overall performance is more competetive than smaller ones. InternVL3-14B achieves the highest overall score (82.40), followed closely by Qwen2.5-14B-Instruct (82.04). Models from the Qwen and InternVL families exhibit balanced capabilities across both customer- and merchant-oriented scenarios, reflecting their strong instruction alignment and task adaptability. 
In particular, InternVL3-14B benefits from extended training data that emphasize tool use, long-context reasoning and creative writing \cite{zhu2025internvl3exploringadvancedtraining}. It also employs Mixed Preference Optimization \cite{mpo}, which leverages additional supervision to better align model responses with ground-truth distributions. This dual-supervision strategy significantly enhances its reasoning and decision-making abilities across diverse e-commerce tasks.
In contrast, Llama3.2-11B and MiniCPM-8B show noticeable performance drops in reasoning-intensive or decision-making tasks such as Scenario Route and Solution Decision, suggesting that these models may lack domain-specific optimization for structured reasoning in e-commerce contexts \cite{minicpm4}. Notably, Qwen3-8B and Qwen3-14B fall behind Qwen3-4B-Instruct. This underscores the impact of post-training quality: Qwen3-4B-Instruct benefits from more recent instruction tuning and alignment optimization \cite{yang2025qwen3technicalreport}, which significantly enhance its task-following and dialogue understanding capabilities in e-commerce scenarios.

\subsubsection{Open-source Models larger than 30B}
Finally, we evaluate the largest models with over 30B parameters. In this group, Qwen3-235B-A22B-Instruct achieves the best overall performance (85.61), likely benefiting from both its large model capacity and the high-quality, diverse-coverage curated fine-tuning samples generated by DeepSeek-R1.
Across these models, we also observe that more recent instruction-tuned variants such as Qwen3-30B-A3B-Instruct and Qwen3-235B-A22B-Instruct consistently outperform their earlier counterparts, indicating the importance of recent instruction tuning that enhance reasoning and task-following abilities \cite{qwen2025qwen25technicalreport}. While most models excel in classification-style tasks like Intent Recognition, their performance in Solution Decision tasks remains unstable, possibly due to the task’s reliance on reasoning over multiple context turns.
Deepseek V3 performs better than Deepseek R1, indicating V3's stronger generalization ability to e-commerce scenarios \cite{deepseekai2025deepseekv3technicalreport}.
Regarding merchant-oriented tasks, dots.llm1.inst demonstrates strong performance in Attitude Classification, likely due to its training on fine-grained dialogue understanding and sentiment detection \cite{huo2025dotsllm1technicalreport}. 


These results can reveal several issues that are often overlooked in standard evaluations. The deficiencies found in different models highlight that even extremely large models can have stage-specific weaknesses that are not apparent when only measuring overall task success. Our stage-wise benchmark is crucial for uncovering these gaps, providing fine-grained insights into where models excel or fail, and guiding targeted improvements for real-world e-commerce applications.

\subsection{Stage-wise and Side-wise Comparison}

\begin{figure}[t]
  \centering
  \includegraphics[width=0.49\textwidth]{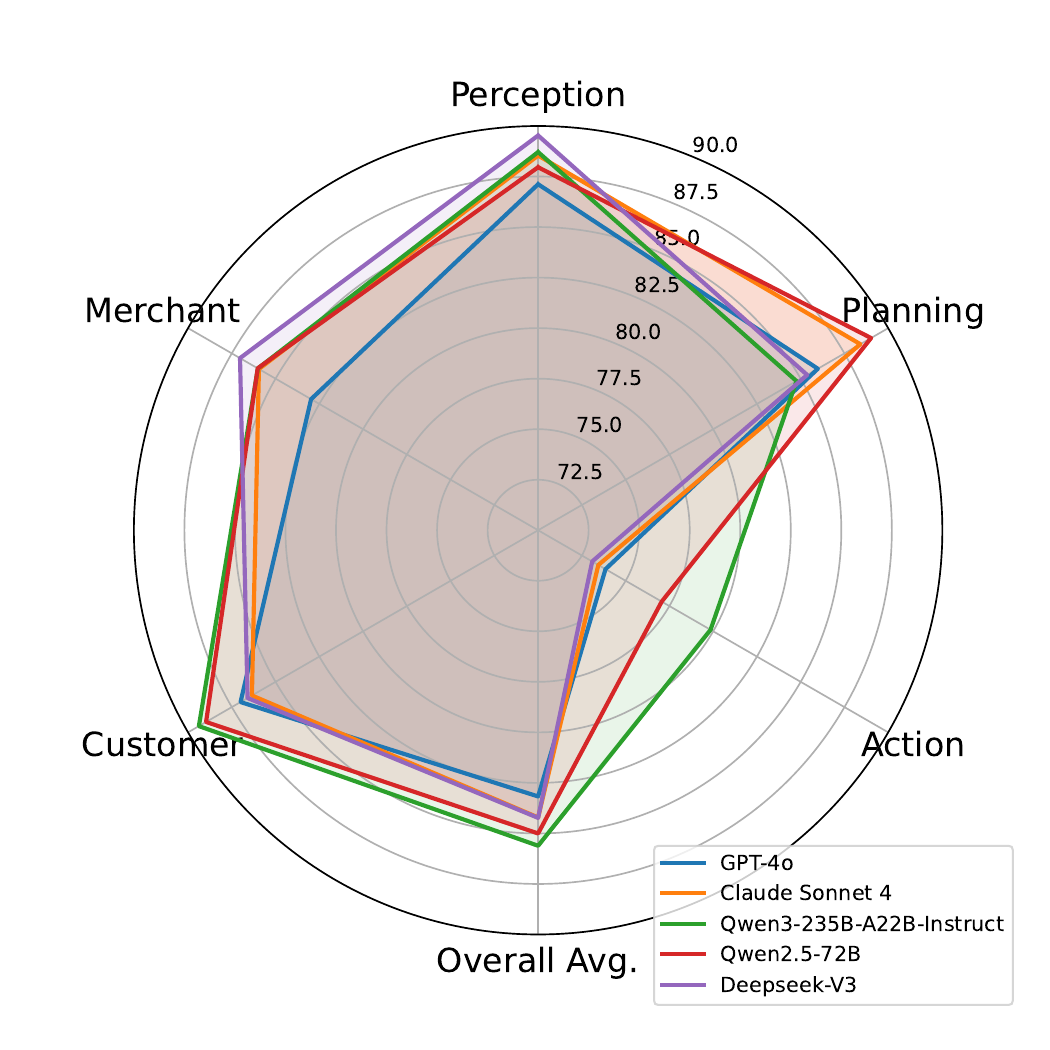}
  \caption{Stage-wise and orientation-wise comparison of LLM-based agents.}
  \label{fig2}

\end{figure}

To better analyze the effect of stage-wise and orientation-wise evaluation, we choose and compare several models that exhibit strong overall performance including GPT4o, Claude-sonnet-4, Qwen3-235B-A22B-Instruct, Qwen2.5-72B-Instruct and Deepseek-V3. We illustrate the stage-wise and orientation-wise performance of them in Figure~\ref{fig2}.

Across the different stages, all models maintain high performance on Perception, while their performance in other stages varies more noticeably. For example, Qwen2.5-72B-Instruct and Claude Sonnet 4 excel in Planning, benefiting from the enhanced multi-step reasoning and the ability to formulate coherent plans \cite{qwen2025qwen25technicalreport, Claude4}. Whereas Qwen3-235B-A22B-Instruct stands out in Action, likely owing to its markedly better alignment with user preferences in open-ended tasks \cite{yang2025qwen3technicalreport}.

As for orientation-wise performance, Qwen3-235B-A22B-Instruct and Qwen2.5-72B-Instruct show strong results on customer-oriented tasks, while Deepseek-V3 exhibits the best performance in merchant-oriented tasks, likely because it has been specifically optimized for task-specific alignment, including structured reasoning and scenario planning. Notably, GPT-4o, previously one of the strongest LLMs, shows relative weakness on merchant-oriented tasks.

These observations highlight that overall average scores alone are insufficient to fully understand model behavior, as they can obscure critical weaknesses at specific stages or on certain sides. Our experiments reveal that \textbf{no single model consistently excels across all tasks, stages, or orientations}.
Our benchmark therefore provides a more granular evaluation framework that captures the multi-dimensional capabilities of LLM-based agents in e-commerce scenarios. By breaking down model performance into Perception, Planning, Action, customer- and merchant-oriented dimensions, it allows researchers and practitioners to identify strengths and weaknesses more precisely and better guide model development and deployment, offering valuable insights for practical applications in both academia and industry.

\section{Conclusion}
In this work, we introduce a comprehensive stage-wise benchmark for evaluating agent-capable LLMs in e-commerce scenarios for both customer and merchant orientations with real-world e-commerce data. Specifically, we divide the reasoning process into Perception, Planning and Action. Our experiments reveal substantial differences in performance across model families, sizes, and instruction-tuning strategies. While most models perform well on Perception and classification-style tasks, significant variation remains in Planning, Action, and merchant-oriented tasks. We observe that no single model can excel across all tasks, underscoring the importance of a granular evaluation framework. Our benchmark enables researchers and practitioners to uncover model strengths and weaknesses that are hidden in success rates, providing actionable insights for model development and deployment.

\section*{Limitations}
Although \Name~decomposes e-commerce reasoning into Perception, Planning, and Action stages, these stages are evaluated through separate, stage-specific tasks. As a result, the benchmark does not capture error propagation across stages, which may occur in real-world deployments. Our benchmark focuses on representative but finite e-commerce scenarios, such as product inquiry, recommendation, complaint handling, and refund decisions. While these tasks cover common real-world interactions, they do not fully encompass all e-commerce domains, leaving room for future expansion.

\section*{Ethics Statement}
\Name~is constructed with careful and strict filtering to avoid introducing potential ethical concerns. Models for evaluation and automatic metrics are used in accordance with their respective licenses. We use OpenAI’s GPT-5 model for minor language editing and grammar polishing. The model is used to improve clarity and conciseness of writing. All technical content, data analysis, and conclusions are produced and validated by the authors.

\bibliography{custom}

\appendix
\begin{figure*}[t]
  \centering
  \includegraphics[width=0.9\textwidth]{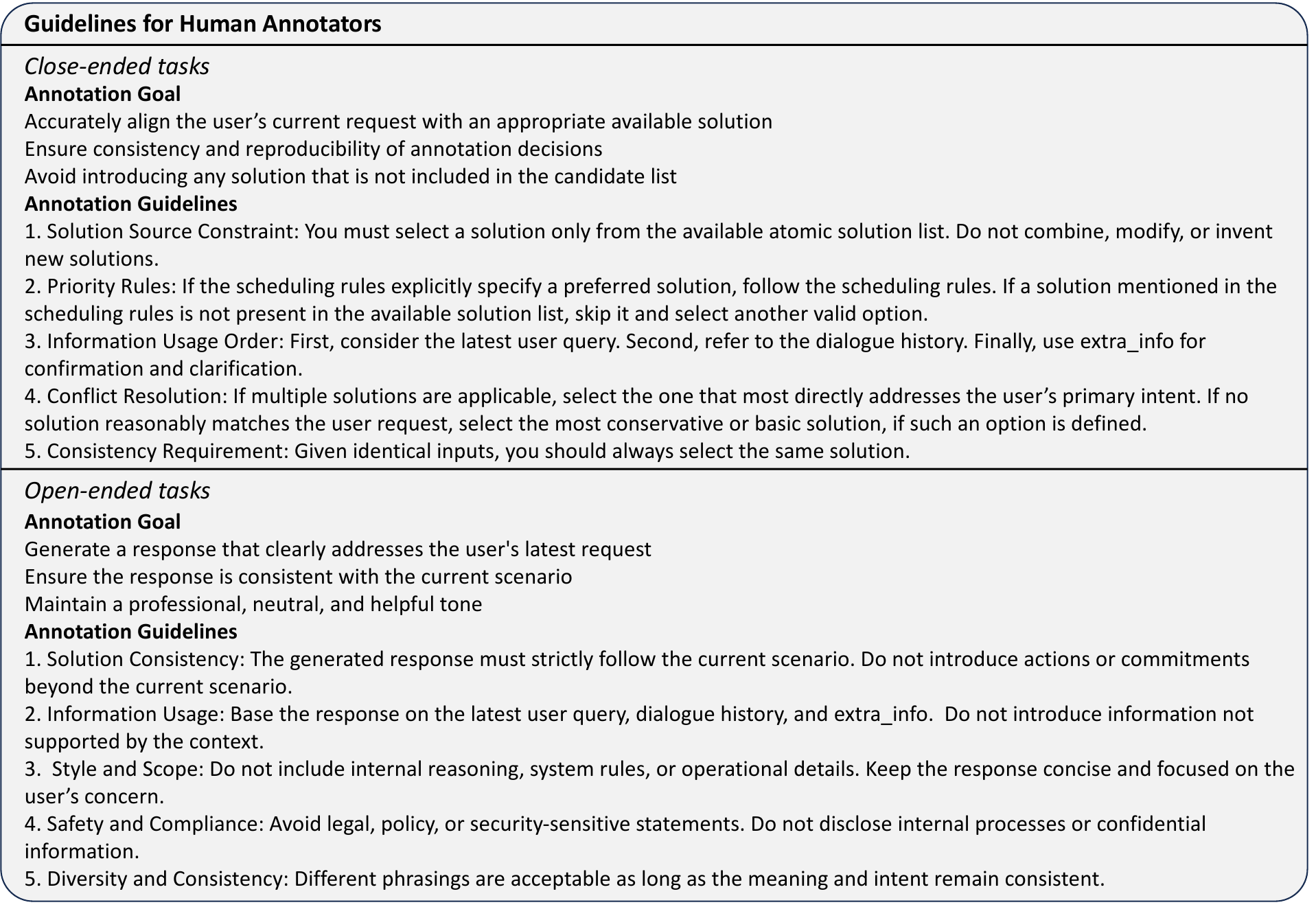}
  \caption{Guidelines for human annotators.}
  \label{guideline}

\end{figure*}
\section{Appendix}
\subsection{Prompts for Dataset Construction}
\label{prompts}
\begin{figure*}[t]
  \centering
  \includegraphics[width=0.9\textwidth]{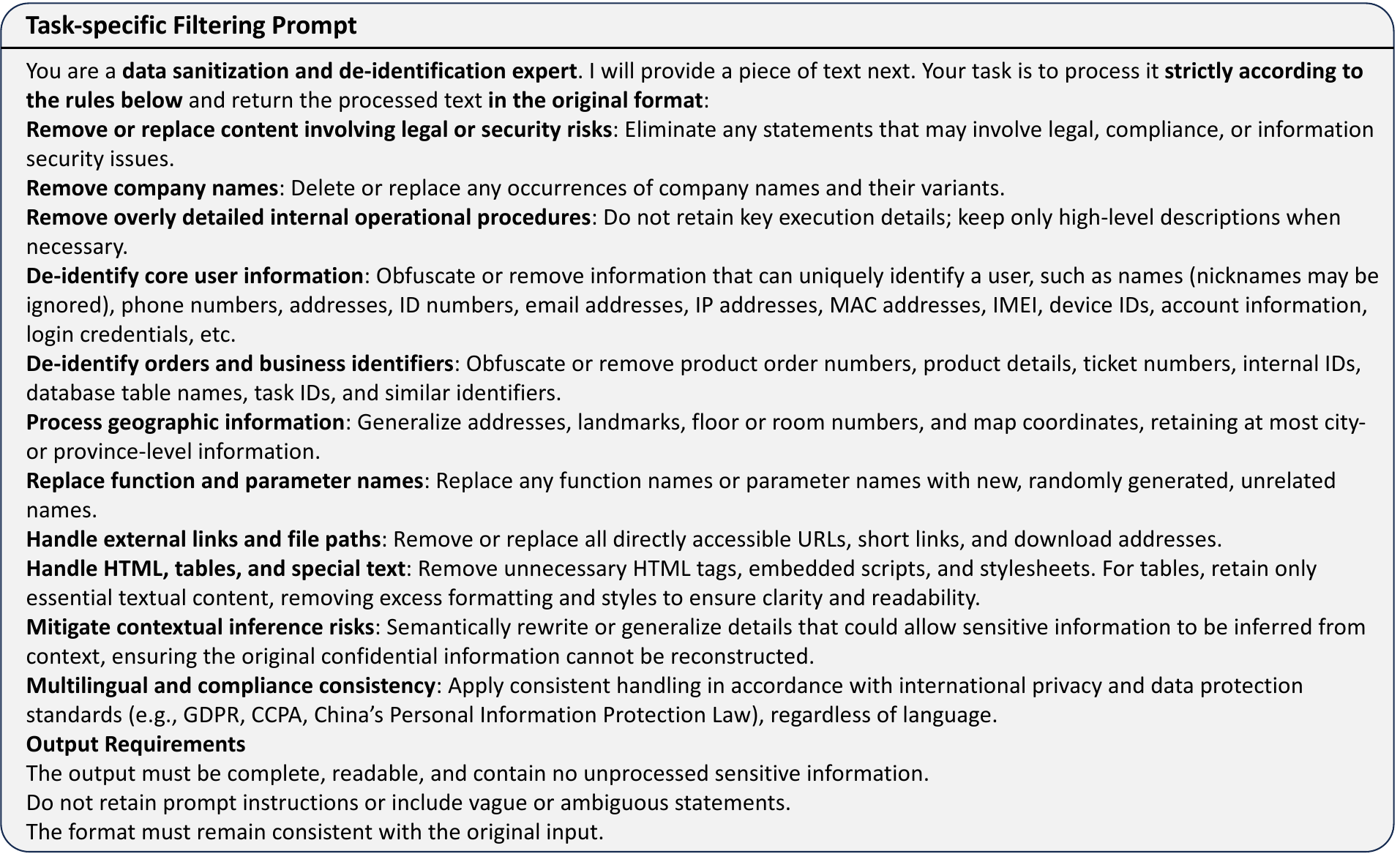}
  \caption{Task-specific filtering prompt for removing sentitive information.}
  \label{task_specific_prompt}

\end{figure*}
\begin{figure*}[t]
  \centering
  \includegraphics[width=0.9\textwidth]{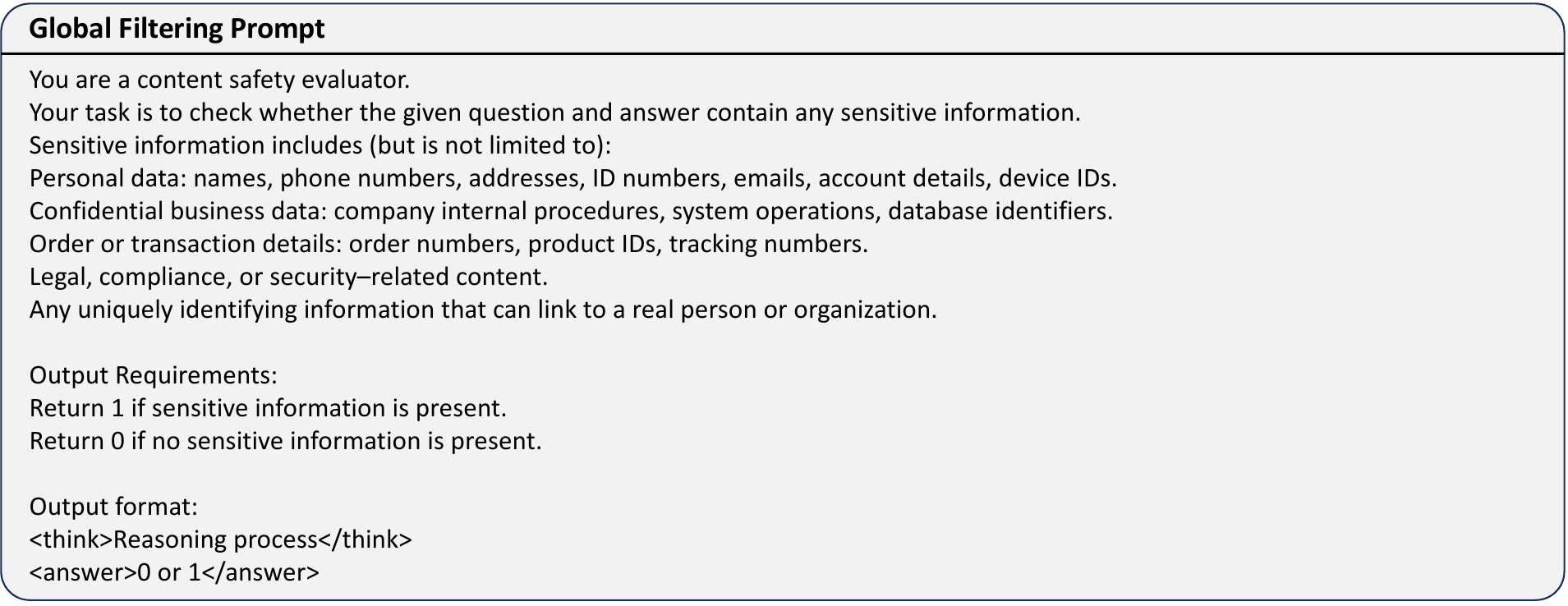}
  \caption{Global filtering prompt to ensure that no sensitive or risky content is involved. We only preserve data labeled as non-sensitive.}
  \label{global_filter_prompt}

\end{figure*}
\begin{figure*}[t]
  \centering
  \includegraphics[width=0.9\textwidth]{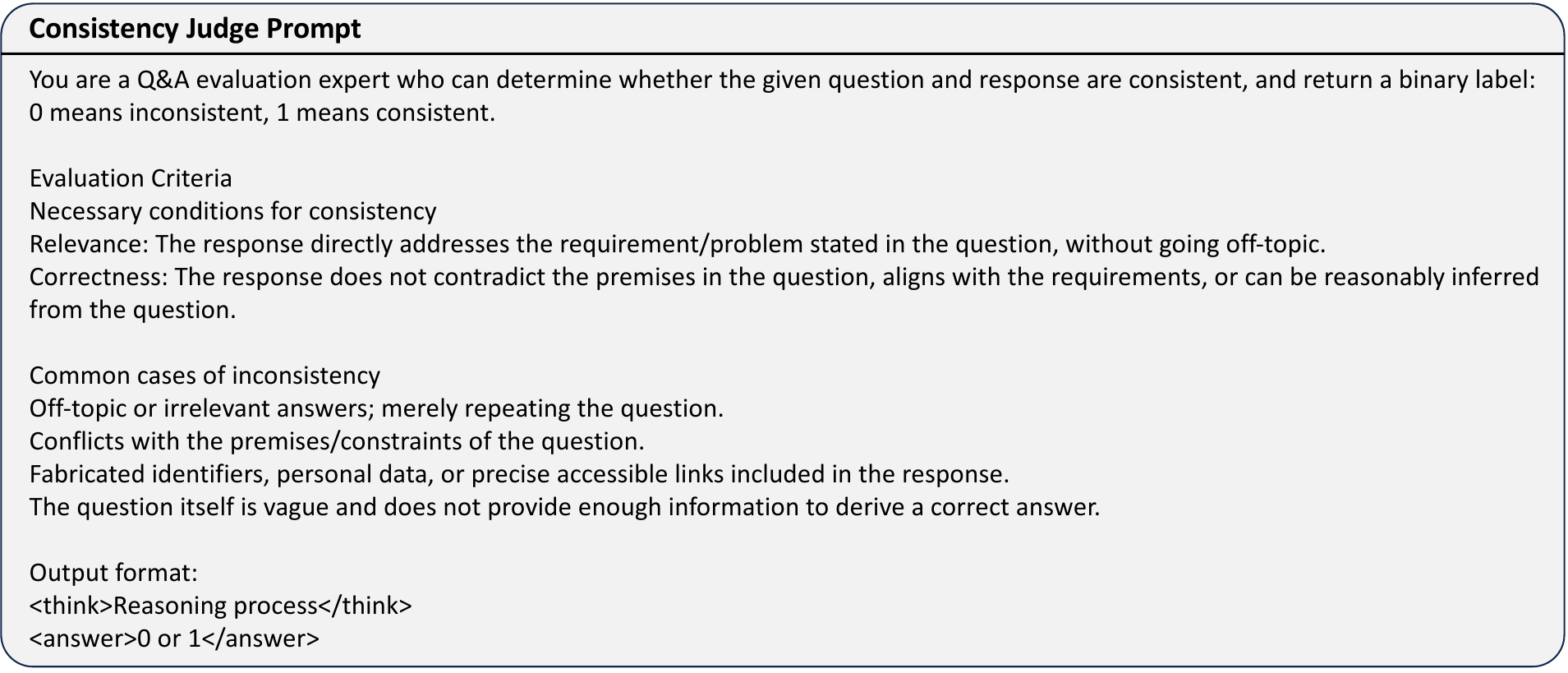}
  \caption{Prompt for consistency judgment. We only preserve data labeled as consistent.}
  \label{consistency_prompt}

\end{figure*}
\begin{figure*}[t]
  \centering
  \includegraphics[width=0.9\textwidth]{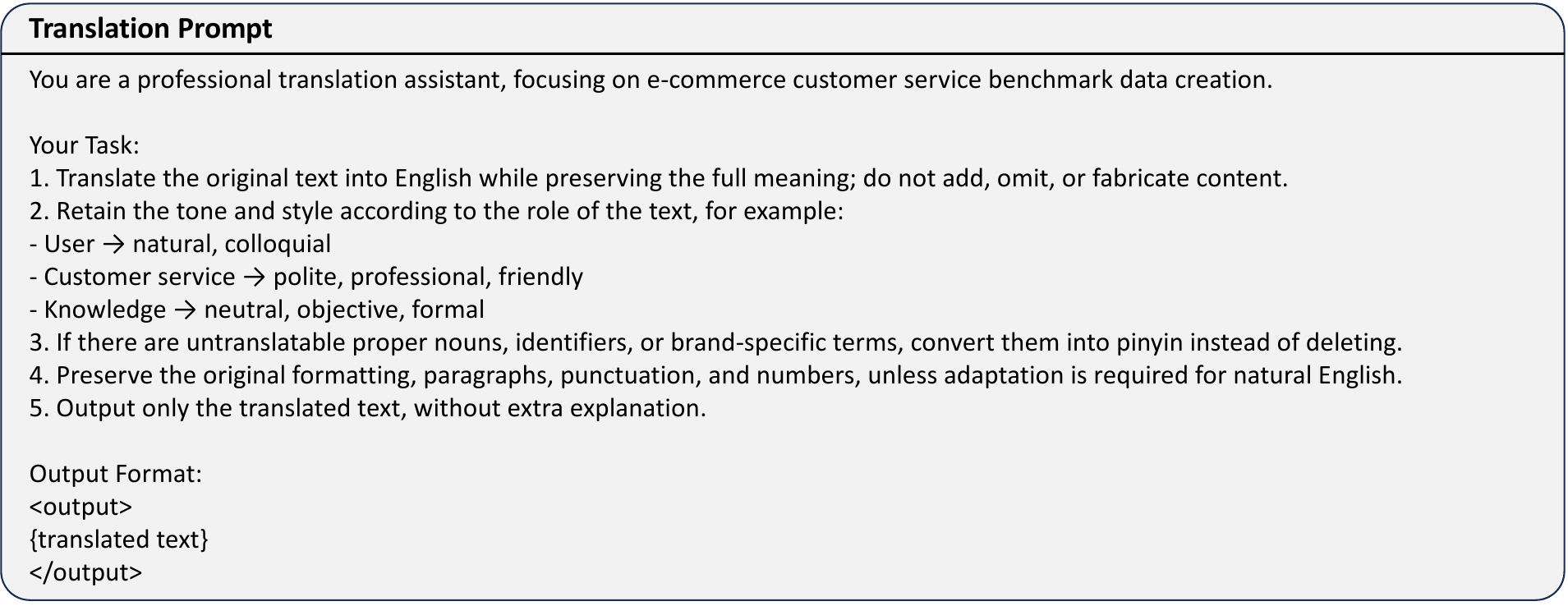}
  \caption{Translation prompt for our data. We translate them from Chinese to English.}
  \label{translation_prompt}

\end{figure*}
We present detailed prompts used in the construction of \Name in this section. Task-specific filtering prompt is presented in Figure~\ref{task_specific_prompt}, while global filtering prompt is provided in Figure~\ref{global_filter_prompt}. Figure~\ref{consistency_prompt} presents the prompt we used for consistency judgment and Figure~\ref{translation_prompt} presents the prompt for translation.

\subsection{Examples for Tasks}
\label{examples}
\begin{figure*}[t]
  \centering
  \includegraphics[width=0.9\textwidth]{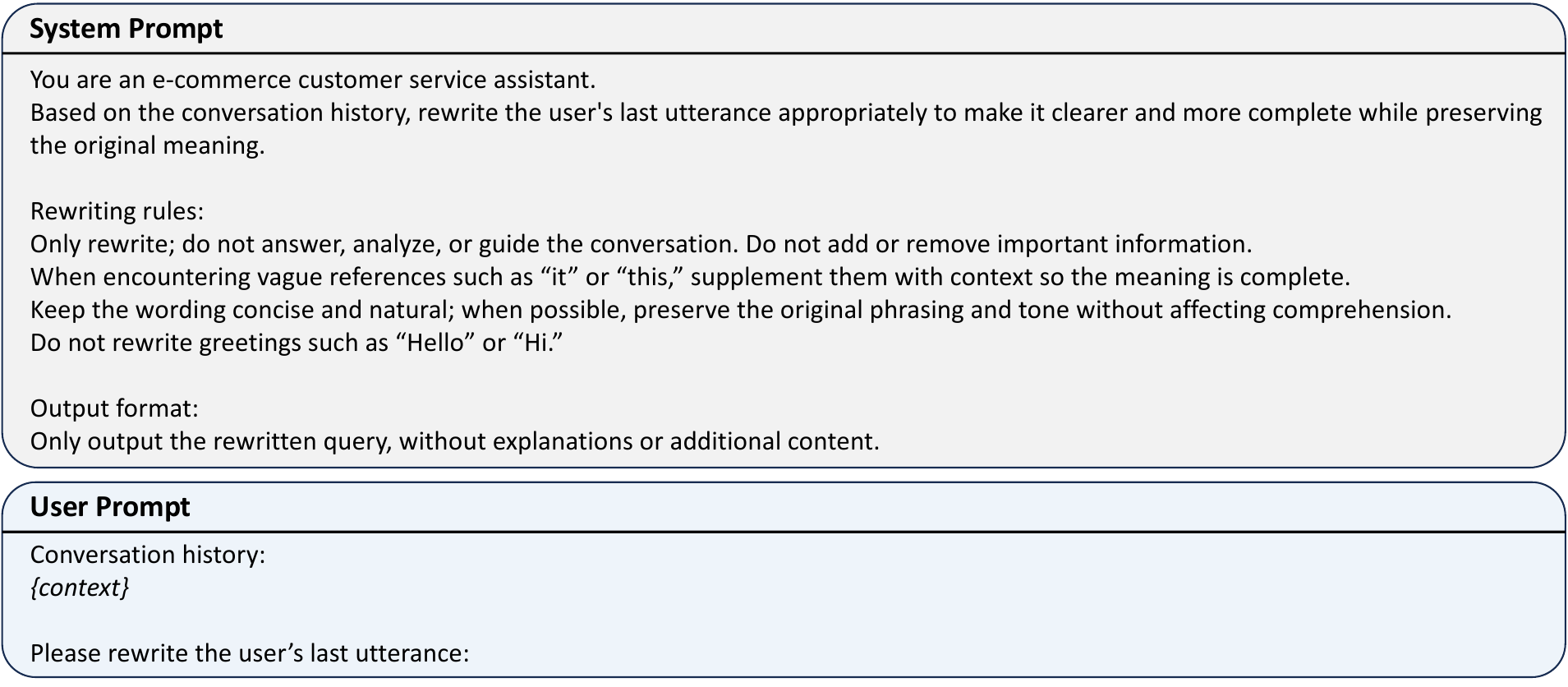}
  \caption{Example for Query Rewrite task (Customer-oriented).}
  \label{query_rewrite}

\end{figure*}
\begin{figure*}[t]
  \centering
  \includegraphics[width=0.9\textwidth]{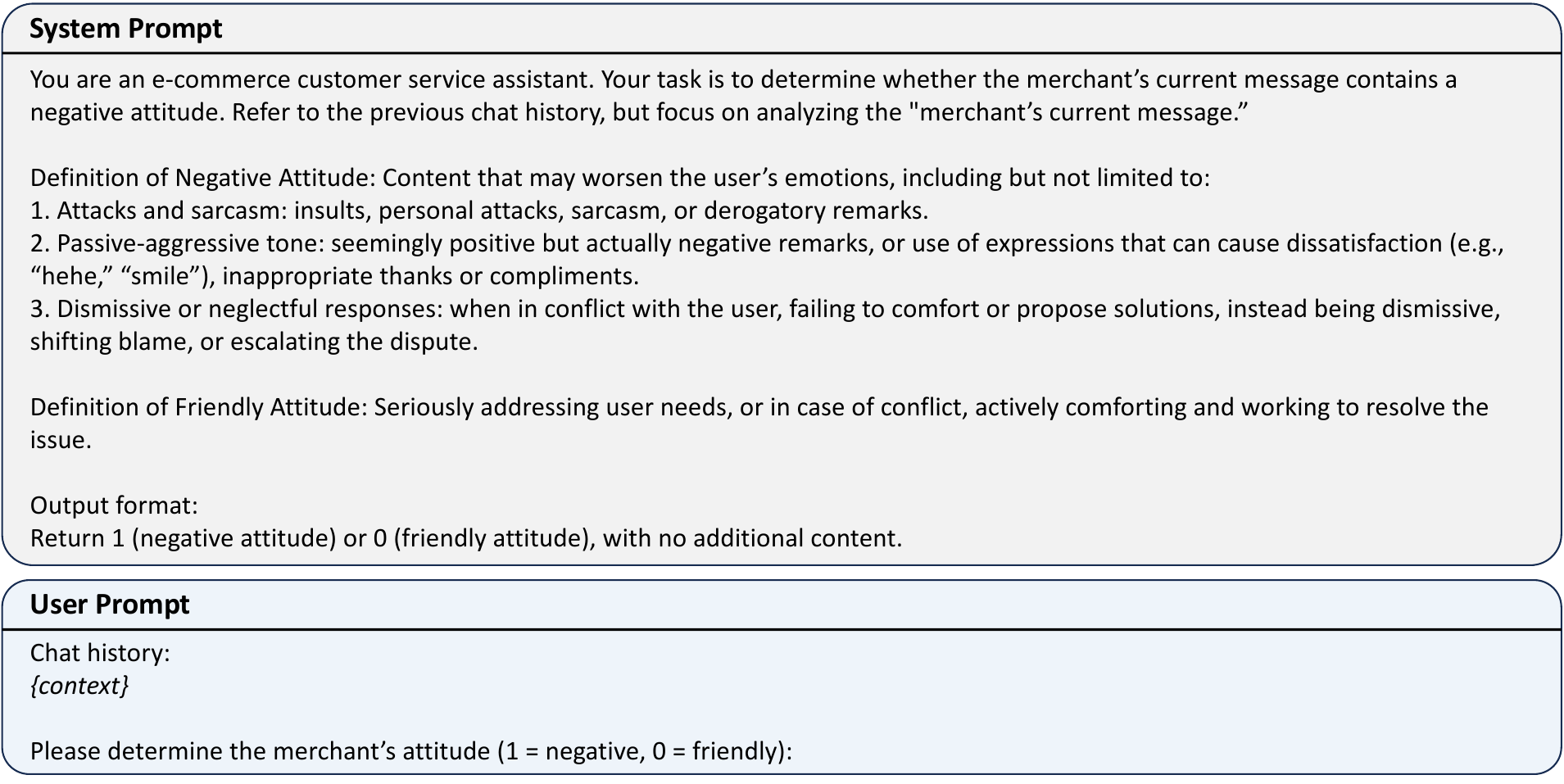}
  \caption{Example for Attitude Classification task (Merchant-oriented).}
  \label{attitude_cls}

\end{figure*}
\begin{figure*}[t]
  \centering
  \includegraphics[width=0.9\textwidth]{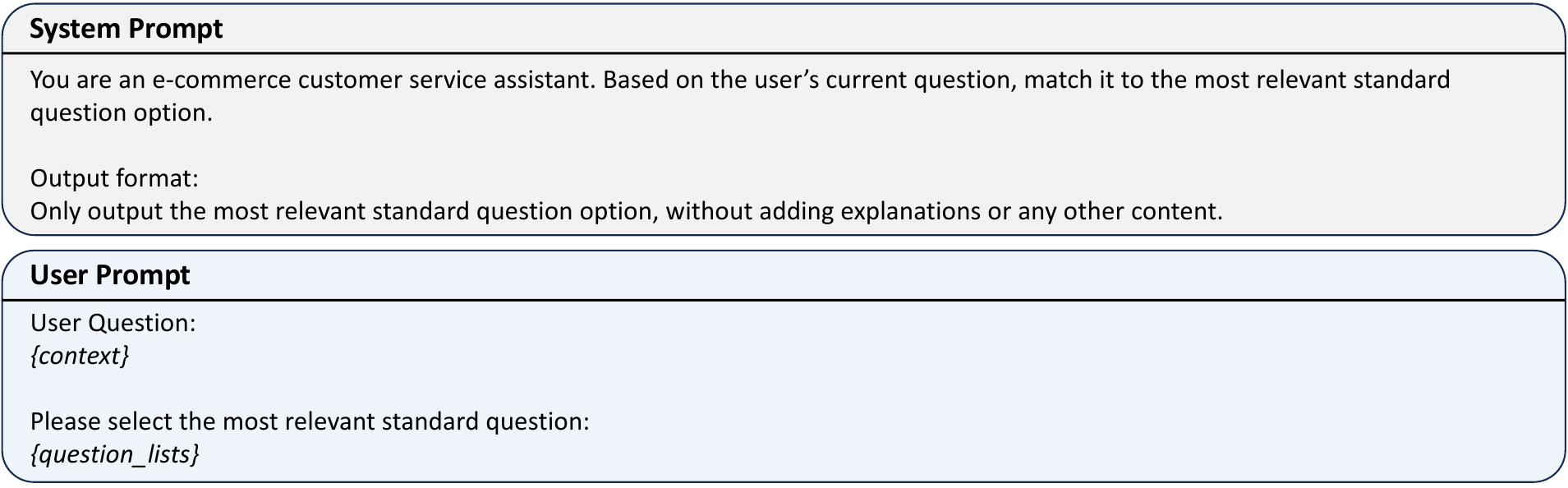}
  \caption{Example for Query Match task (Customer-oriented).}
  \label{query_match}

\end{figure*}
\begin{figure*}[t]
  \centering
  \includegraphics[width=0.9\textwidth]{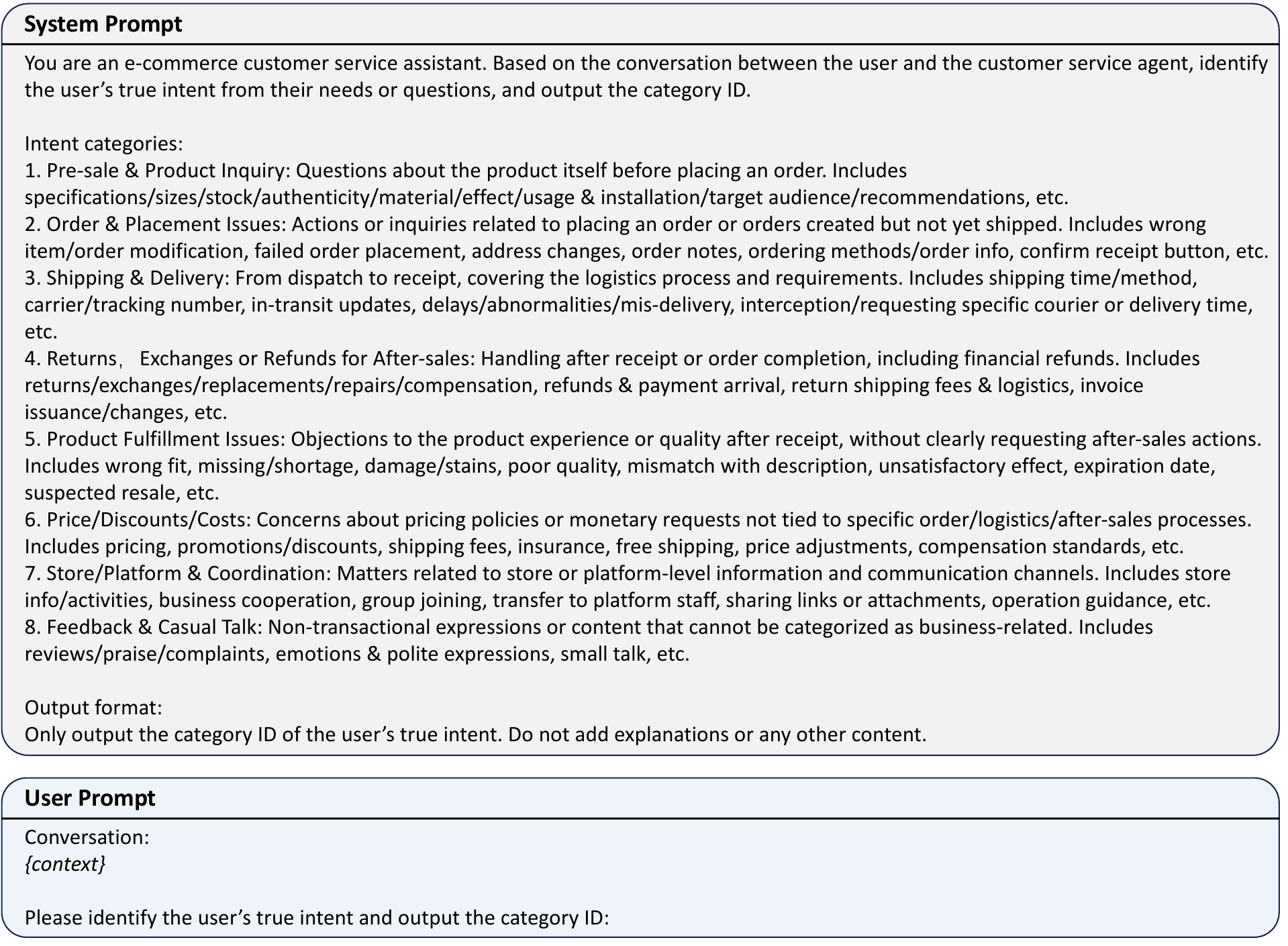}
  \caption{Example for Intent Recognition task (Customer-oriented).}
  \label{intent_reco}

\end{figure*}
\begin{figure*}[t]
  \centering
  \includegraphics[width=0.9\textwidth]{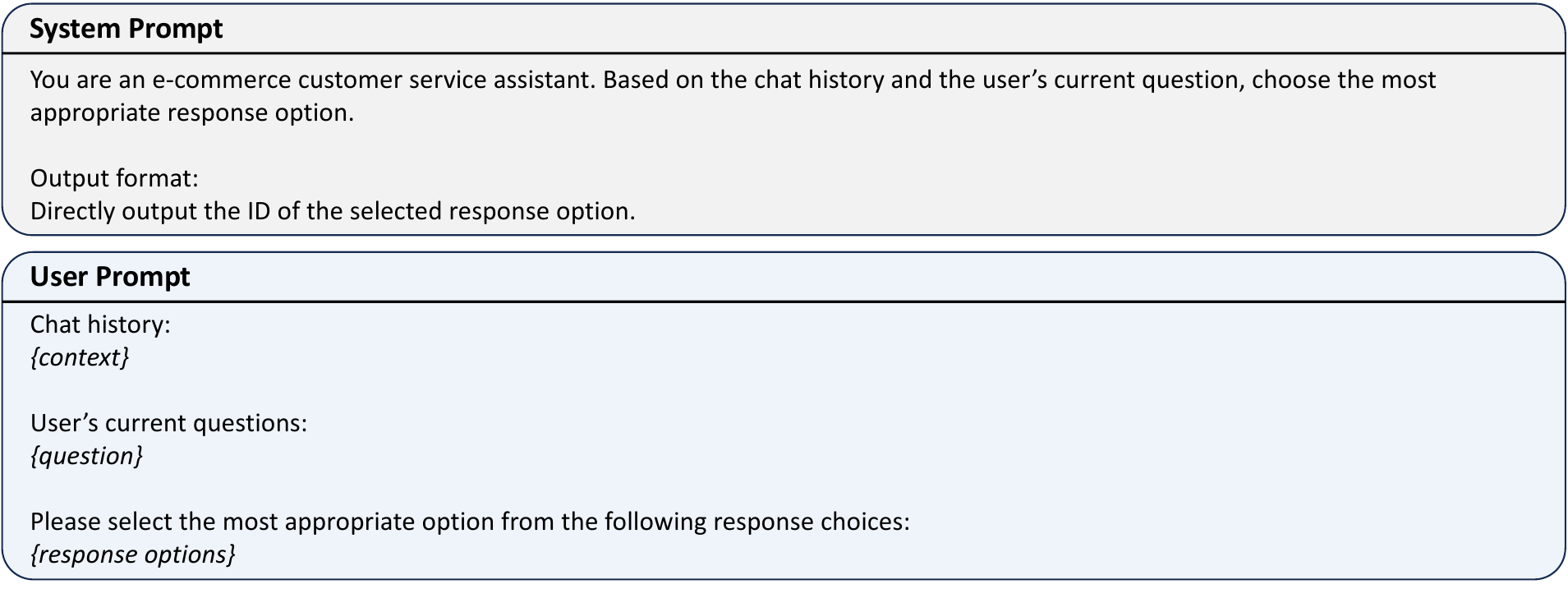}
  \caption{Example for Solution Decision task (Customer-oriented).}
  \label{solution_decision}

\end{figure*}
\begin{figure*}[t]
  \centering
  \includegraphics[width=0.9\textwidth]{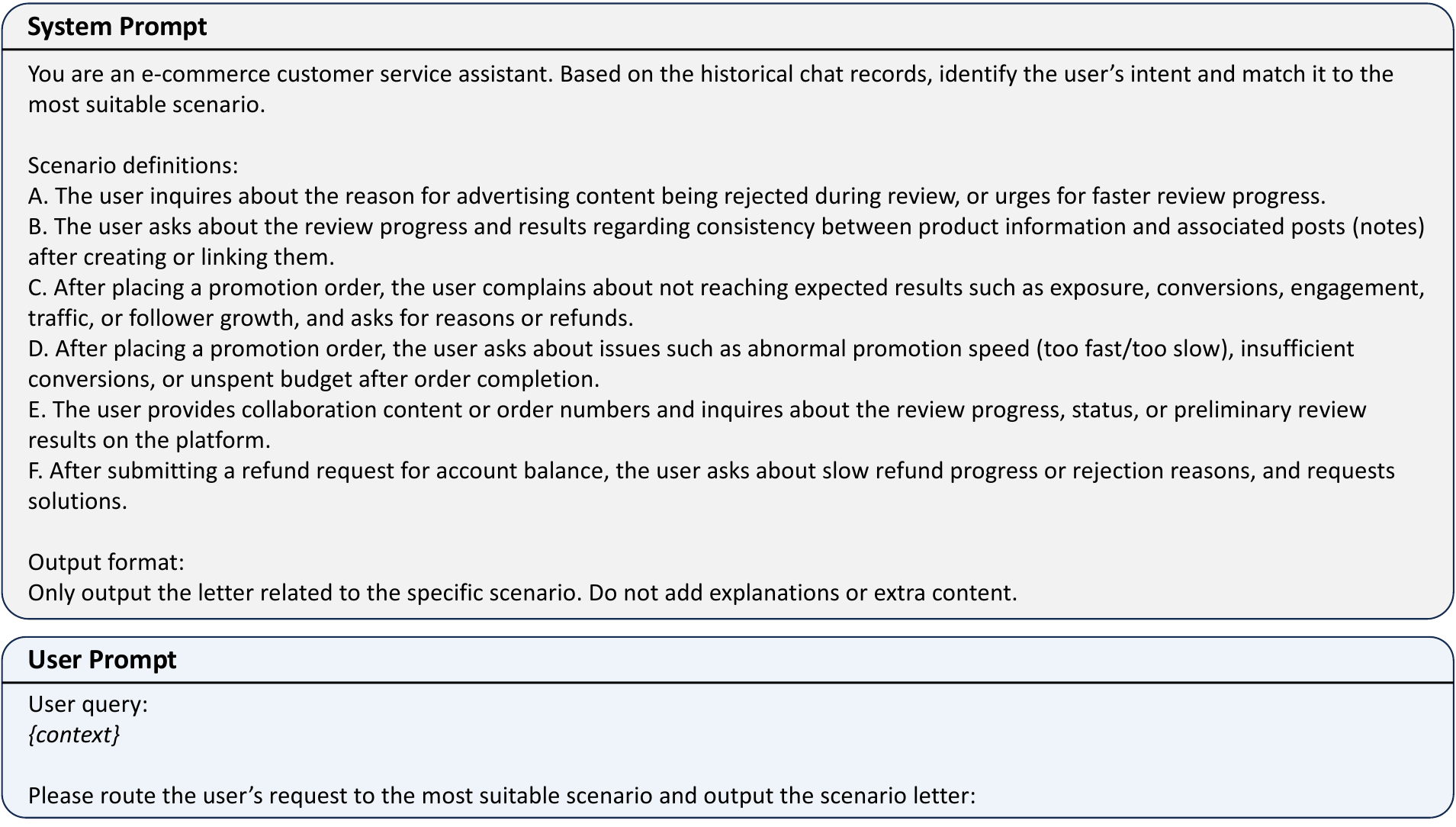}
  \caption{Example for Scenario Route task (Merchant-oriented).}
  \label{scenario_route}

\end{figure*}
\begin{figure*}[t]
  \centering
  \includegraphics[width=0.9\textwidth]{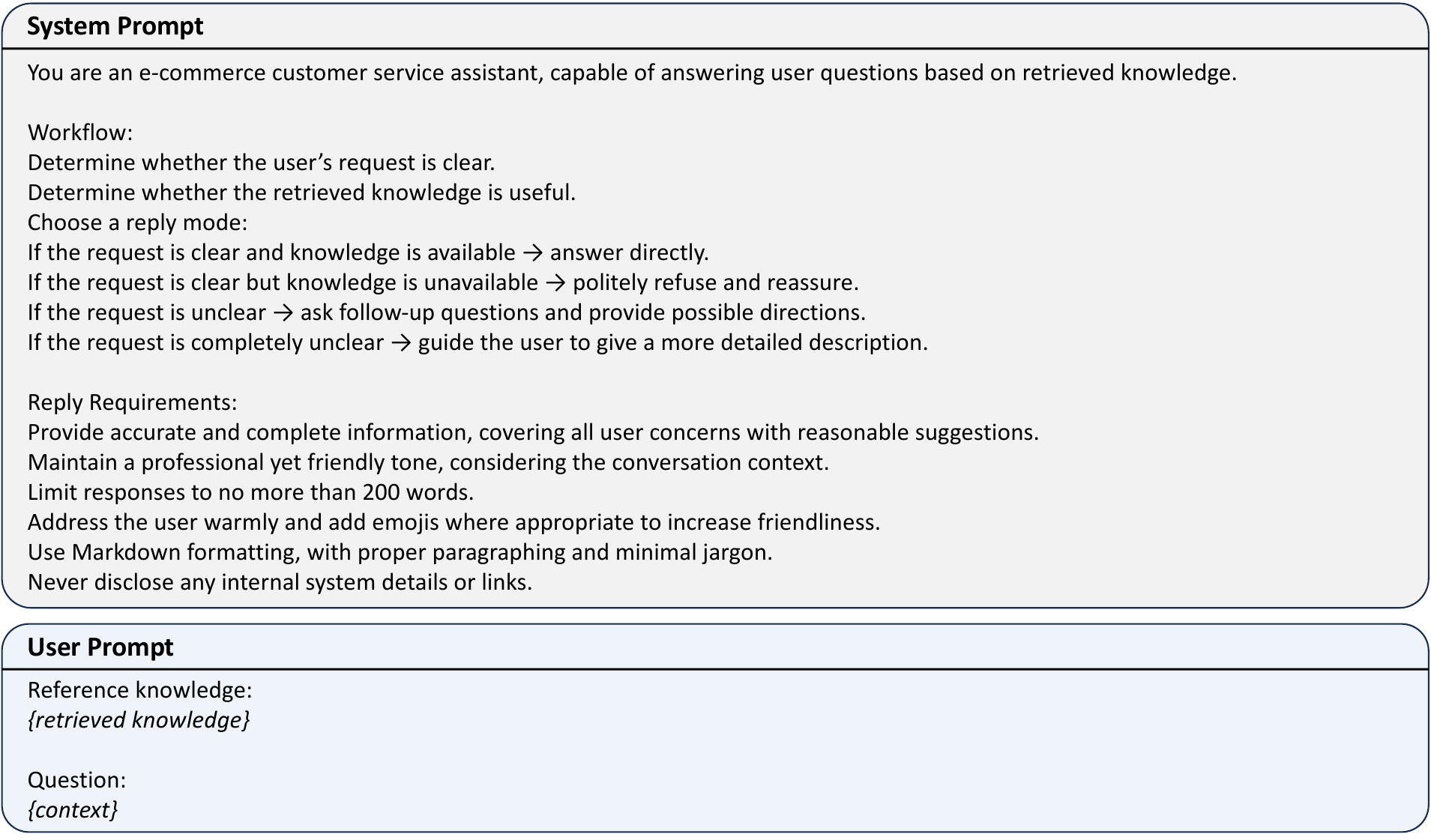}
  \caption{Example for RAG-QA task (Customer- and Merchant-oriented).}
  \label{ragqa}

\end{figure*}
We present examples for our seven tasks in Figure~\ref{query_rewrite}, \ref{attitude_cls}, \ref{query_match}, \ref{intent_reco}, \ref{solution_decision}, \ref{scenario_route}, and \ref{ragqa}, separately.

\subsection{Human Annotators}
All participants involved in the annotation and evaluation process are employees of our company with relevant experience in e-commerce operations. They are assigned with the tasks as part of their regular work responsibilities, and no additional payment is provided beyond their standard compensation. Given their professional expertise and familiarity with e-commerce scenarios, they are well-qualified to perform the tasks reliably.

\end{document}